\documentclass[journal]{IEEEtran}

\usepackage{algorithm}
\usepackage{algorithmicx}
\usepackage[noend]{algpseudocode} 

\usepackage{cite}
\usepackage{acro}
\usepackage{graphicx}
\usepackage{float}
\usepackage{xcolor}
\usepackage{amsmath}
\usepackage{multirow}
\usepackage{multicol}
\usepackage{wrapfig}
\usepackage{booktabs}   
\usepackage{subcaption}
\usepackage{wrapfig}
\usepackage{float}

\DeclareAcronym{FCN}{
short=FCN,
long=fully convolutional network,
}

\DeclareAcronym{CNN}{
short=CNN,
long=convolutional neural network,
}

\DeclareAcronym{MOT}{
short=MOT,
long=Multiple object tracking,
}

\DeclareAcronym{SOT}{
short=SOT,
long=single object tracking,
}

\DeclareAcronym{LSTM}{
short=LSTM,
long=long short-term memory,
}

\DeclareAcronym{DPA}{
short=DPA,
long=Dual-path association,
}

\DeclareAcronym{Map3D}{
short=Map3D,
long=Multi-object Association for Pathology in 3D,
}

\DeclareAcronym{IoU}{
short=IoU,
long=Intersection over Union,
}

\DeclareAcronym{SIFT}{
short=SIFT,
long=scale-invariant feature transform,
}

\DeclareAcronym{SG}{
short=SG,
long=SuperGlue,
}

\DeclareAcronym{QaWS}{
short= QaWS,
long=quality-aware whole series ,
}

\DeclareAcronym{QA}{
short= QA,
long=quality assurance,
}

\DeclareAcronym{ANTs}{
short= ANTs,
long=advanced normalization tools,
}

\begin{document}
%
\title{Semantic-Aware Contrastive Learning for Multi-object Medical Image Segmentation}
%
%
%

\author{Ho Hin Lee,
        Yucheng Tang,
        Qi Yang,
        Xin Yu,
        Leon Y. Cai,
        Lucas W. Remedios,
        Bennett A. Landman,~\IEEEmembership{Senior Member,~IEEE},
        Yuankai~Huo,~\IEEEmembership{Senior Member,~IEEE}
\thanks{This research is supported by NIH Common Fund and National Institute of Diabetes, Digestive and Kidney Diseases U54DK120058, NSF CAREER 1452485, NIH 2R01EB006136, NIH 1R01EB017230, and NIH RO1NS09529. ImageVU and RD are supported by the VICTR CTSA award (ULTR000445 from NCATS/NIH). We gratefully acknowledge the support of NVIDIA Corportaion with the donation of the Titan X Pasacl GPU usage.}
\thanks{Ho Hin Lee, Qi Yang, Xin Yu, Lucas W. Remedios, Shunxing Bao, Yuankai Huo are with the Computer Science Department, Vanderbilt University, Nashville TN 37235}
\thanks{Yucheng Tang, Bennett A. Landman are with the Electrical and Computer Engineering Department, Vanderbilt University, Nashville TN 37235}
\thanks{Leon Y. Cai is with the Biomedical Engineering Department, Vanderbilt University, Nashville TN 37235}
\thanks{Ho Hin Lee is the first author to the manuscript. (Corresponding author: Yuankai Huo, Email: yuankai.huo@vanderbilt.edu)}
}

\markboth{Manuscript pre-print, November~2021}%
{Shell \MakeLowercase{\textit{et al.}}: Bare Demo of IEEEtran.cls for IEEE Journals}

\maketitle

\begin{abstract}
Medical image segmentation, or computing voxel-wise semantic masks, is a fundamental yet challenging task to compute a voxel-level semantic mask. To increase the ability of encoder-decoder neural networks to perform this task across large clinical cohorts, contrastive learning provides an opportunity to stabilize model initialization and enhance encoders without labels. However, multiple target objects (with different semantic meanings) may exist in a single image, which poses a problem for adapting traditional contrastive learning methods from prevalent ``image-level classification'' to ``pixel-level segmentation''. In this paper, we propose a simple semantic-aware contrastive learning approach leveraging attention masks to advance multi-object semantic segmentation. Briefly, we embed different semantic objects to different clusters rather than the traditional image-level embeddings. We evaluate our proposed method on a multi-organ medical image segmentation task with both in-house data and MICCAI Challenge 2015 BTCV datasets. Compared with current state-of-the-art training strategies, our proposed pipeline yields a substantial improvement of 5.53\% and 6.09\% on Dice score for both medical image segmentation cohorts respectively (p-value$<$0.01). The performance of the proposed method is further assessed on natural images via the PASCAL VOC 2012 dataset, and achieves a substantial improvement of 2.75\% on mIoU (p-value$<$0.01). 
\end{abstract}

\begin{IEEEkeywords}
Medical image segmentation, contrastive learning, attention map, query patches
\end{IEEEkeywords}

%
\IEEEpeerreviewmaketitle

\section{Introduction}
\label{sec:introduction}
\IEEEPARstart{C}{contrastive} learning is a variant of self-supervised learning (SSL) that has advanced to learn useful representation for image classification task \cite{chen2020simple}. Traditional contrastive learning approach consists of two primary concepts: 1) the learning process should pull the target image (anchor) and a matching sample as together ``positive pair'', and 2) the learning process should push the anchor from numerous non-matching samples apart (``negative pair''). In self-supervised setting, data augmentation is used to generate positive samples from the anchor, while the negative pairs (dissimilar samples) are generated from the remaining samples of non-matching objects. Previous work demonstrates significant improvement in image-level classification tasks with object-centric images \cite{NEURIPS2020_63c3ddcc,NEURIPS2020_3fe23034,CHEN2021107826}. We posit that the encoding ability of the model can be improved by adapting object-centric representation for contrastive learning. However, there remains a gap regarding contrastive learning for pixel-level semantic segmentation tasks \cite{NEURIPS2020_949686ec,liu2021domain,zhao2021contrastive}. Multiple objects may exist, and it is challenging to adapt multiple semantic representations in a single image without additional guidance. 



\begin{figure*}[htb]
\centering
\includegraphics[width=\textwidth]{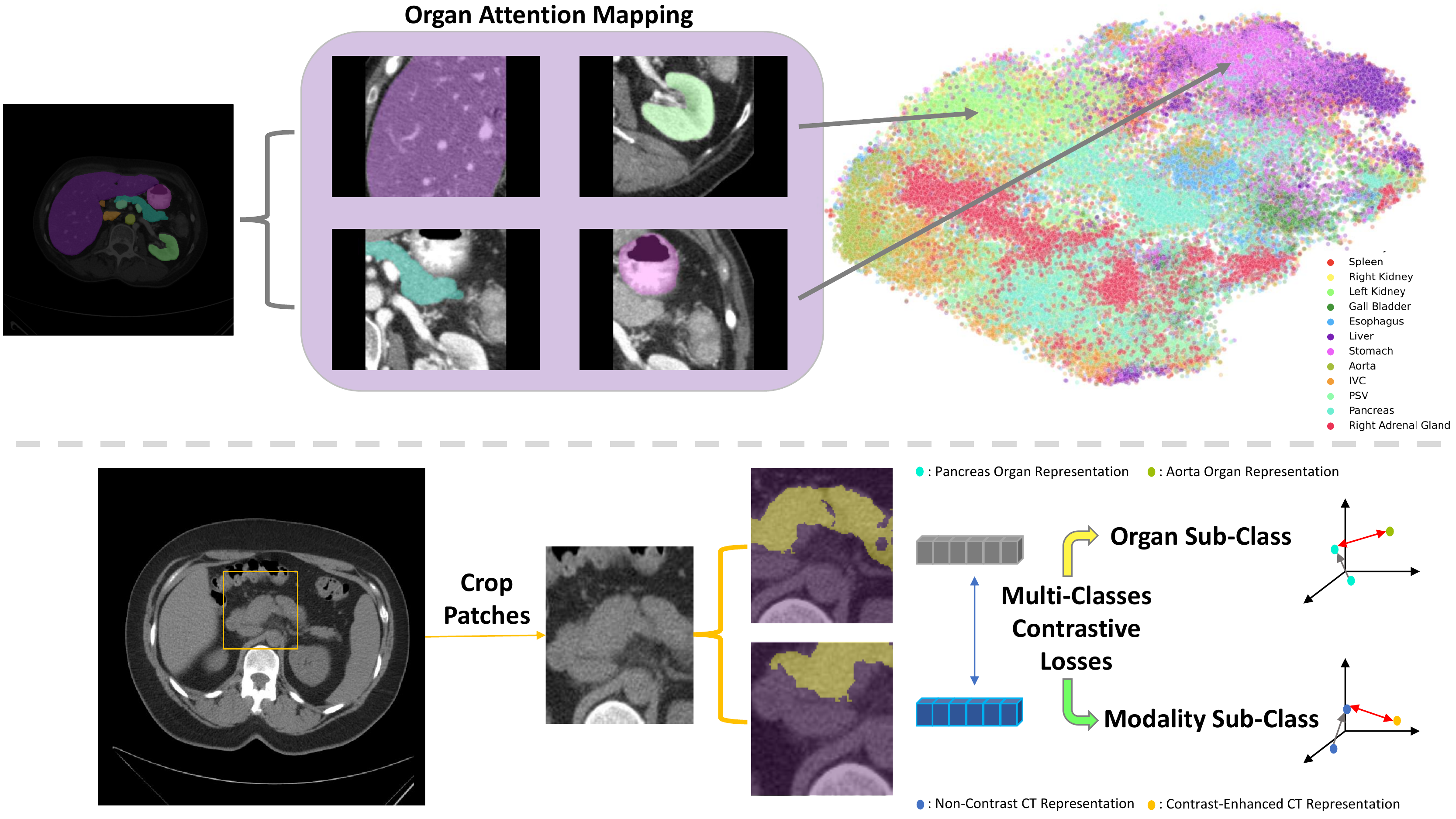}
\caption{With multiple organs located in a single image, organ attention maps  guide their representations into corresponding embeddings and adapt contrastive learning for multi-object segmentation. Categorical information can be used for supervisory context to constrain the separation of clusters (grey arrow: pull the matching representations together, red arrow: push the non-matching representations apart).} \label{problem_figure}
\end{figure*}

In this work, we propose a semantic-aware attention-guided contrastive learning (AGCL) framework to advance multi-object medical image segmentation with contrastive learning. We integrate object-corresponding attention maps as additional input channels to adapt representations into corresponding semantic embeddings (as shown in Fig. 1). To further stabilize the latent space, we propose a multi-class conditional contrastive loss that leverages the number of positives with the same sub-classes. Radiological conditions such as modality and organ information are provided as additional information to constrain the normalized embedding. Fig. 2 provides a visual explanation of our proposed framework. Our contrastive learning
proposed learning strategy AGCL is evaluated with two medical imaging datasets (public contrast-enhanced CT dataset \cite{landman2015miccai}, in-house non-contrast dataset) and one natural imaging dataset (the validation set of PASCAL VOC 2012 \cite{pascal-voc-2012}). The results demonstrate that a consistent improvement is achieved on ResNet-50 and ResNet-101 architectures \cite{he2016deep}. Our main contributions are summarized as below:

\indent1. We propose a semantic-aware contrastive learning framework to advance multi-object pixel-level semantic segmentation. \\
\indent2. We propose a multi-conditional contrastive loss to integrate meta information as an additional constraint. \\
\indent3. We demonstrate that the proposed AGCL generalizes the CT contrast phase variation in each organ and significantly boosts the segmentation performance in an end-to-end architecture. \\

\begin{figure*}[htb]
\centering
\includegraphics[width=\textwidth]{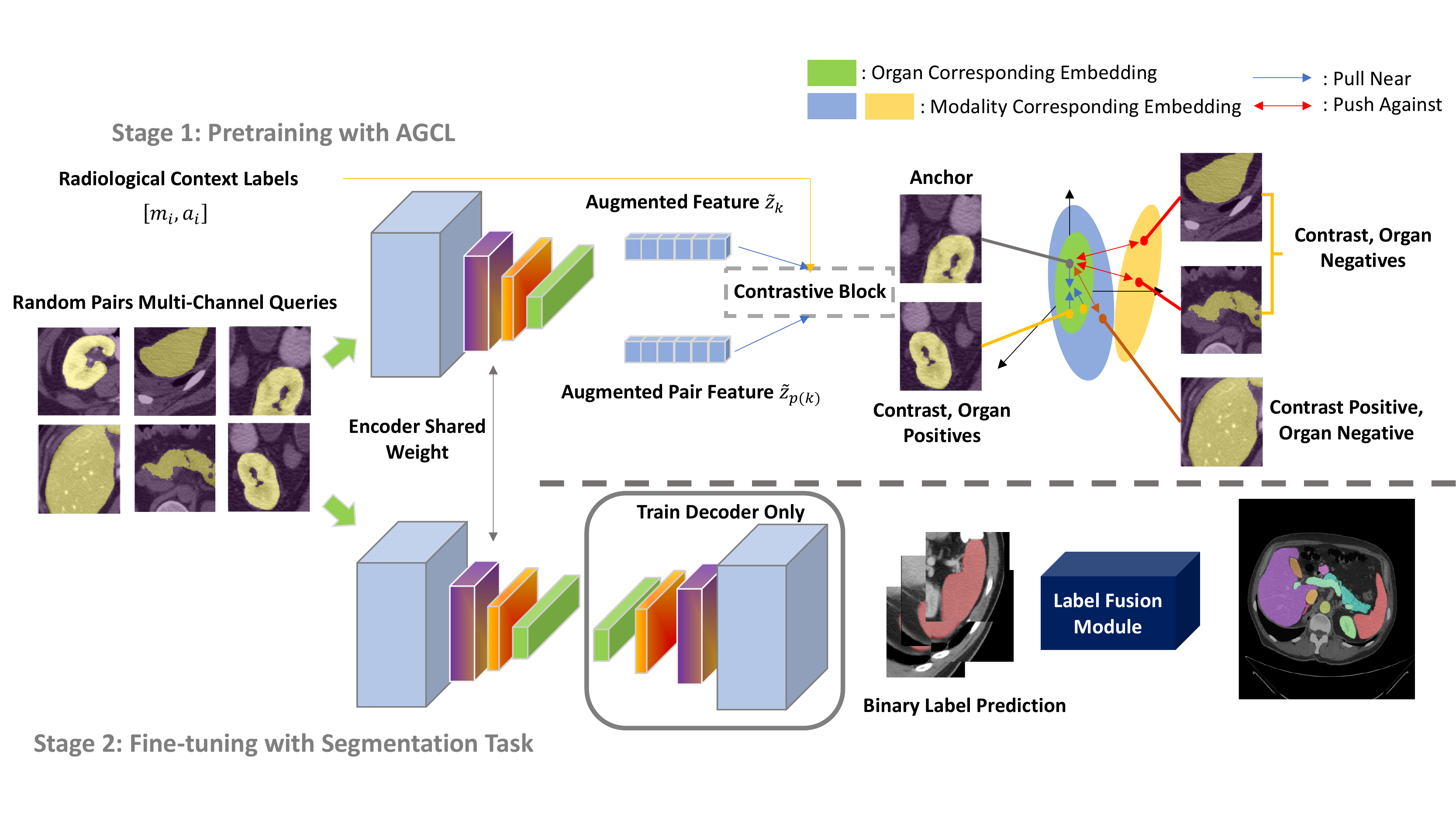}
\caption{A 2D/3D segmentation pipeline (2D for natural image, 3D for medical image) is used to generate attention map for organ localization. 2D organ-corresponding query patches are randomly extracted and concatenated with the regional attention maps as an additional channel to guide embeddings of the organ targets. Data augmented pairs of the attention queries are constrained into corresponding radiological embeddings (such as organs and modalities) with additional label supervision in the proposed contrastive loss. The well-pretrained encoder weights are shared, and the decoder is subsequently trained during the second stage to generate refined segmentations with label fusion.} \label{pipeline_figure}
\end{figure*}

\section{Related Works}
\textbf{Contrastive Learning:} Self-supervised representation learning approaches have recently been proposed to learn useful representation from unlabeled data. Some approaches propose learning embeddings directly in lower-dimensional representation spaces instead of computing a pixel-wise predictive loss~\cite{zhang2017split}. Contrastive learning is one such state-of-the-art methods for self-supervised learning to model the pixel relationships in the latent space~\cite{zhao2020contrastive,chen2020simple}. It employs a loss function to pull latent representations closer together for positive pairs, while pushing them apart for negative pairs. Maximizing mutual information between embeddings has also been proposed as an alternative to extract the similar information between targets \cite{bachman2019learning}. Adapting with memory bank and momentum contrastive approaches have been proposed to increase the batch sizes and generate more dissimilar pairs in a minibatch for contrastive learning \cite{misra2020self}. Additionally, to constrain and stabilize the embedding spaces, class label information has been added to provide additional supervision to stabilize contrastive learning process \cite{khosla2020supervised}. \par

\textbf{Medical Image Segmentation:} Basic approaches to perform medical image segmentation typically involved directly training a deep network model with post-processing techniques \cite{wang2021view}. However, the performance of supervised learning strategies is limited to the quality of voxel-wise labels and the resolution of volumes \cite{gamechi2019automated}. Thus, hierarchical approaches and patch-wise approaches have been proposed to perform segmentation across scales and compute refined segmentations with the resolution variability \cite{roth2018multi,zhu2019multi, tang2021high}. However, multiple models are typically needed to be trained for multiple semantic targets segmentation. Single patch-wise networks have been adapted to perform multi-organs segmentation by adding channels with binary pseudo-segmentation mappings \cite{lee2021rap}. Apart from deep supervised approaches, semi-supervised learning and self-supervised learning have also been explored to adapt unlabeled data in the medical imaging domain. A quality assurance module have been proposed to adapt the segmentation quality as the supervision from unlabeled data \cite{lee2020semi}. Pretext tasks such as colorization, deformation and image rotation, have been used as pre-training features to initialize the segmentation networks \cite{zhou2021models}. Self-supervised context has been explored by predicting the relative patch location and the degree of rotation \cite{bai2019self, zhuang2019self}. Contrastive learning has been used to extract global and local representations for domain-specific MRI images \cite{NEURIPS2020_949686ec}. A contrastive predictive network has been used to summarize the latent vectors in a minibatch and predict the latent representation of adjacent patches \cite{taleb20203d}.

\section{Method}
In this section, we explore the usage of semantic attention queries to adapt contrastive learning for multi-object segmentation and develop the theoretical derivation of a multi-conditional contrastive loss for medical image segmentation with multi-object images in single network architecture.

\subsection{Contrastive Learning Framework for Segmentation} 
Given a set of $N$ randomly sampled 2D patches $P_i = {\{x_i, y_i, c_i\}_{i=1,...,N}}$, where $N$ is the total number of query patches in a minibatch, $x_i$ is the image patch, and $y_i$ and $c_i$ are the corresponding binary label patch and coarse attention patch, respectively. Each $c_i$ is initially concatenated with $x_i$ as a multi-channel input $a_i$ for contrastive learning in the first stage \cite{lee2021rap}. The integration of attention map aims to minimize the possibility of extracting features from the neighboring organs. Additionally, categorical context (such as organ sub-class and modality sub-class) can be classified and further use as supervisory label to leverage the stability of learning independent embeddings. The framework mainly consists of:

\begin{figure*}[htb]
\centering
\includegraphics[width=\textwidth]{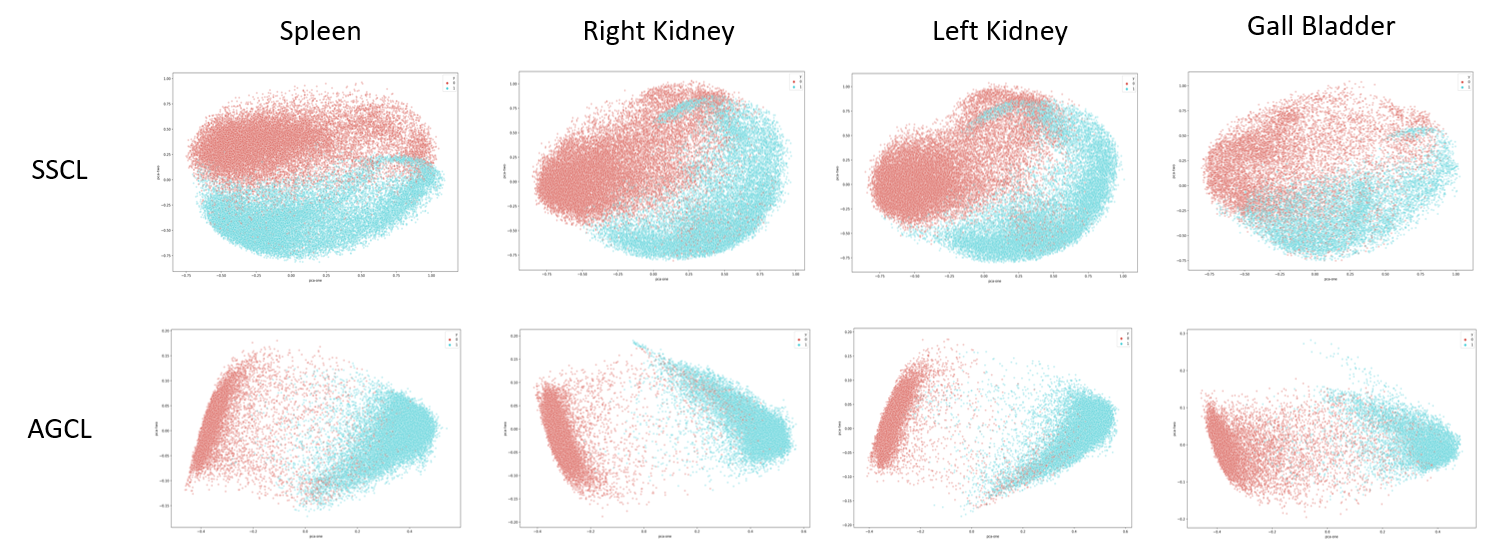}
\caption{The latent distributions of four randomly selected organs plotted with their corresponding modality (Blue: contrast-enhanced phase CT, red: non-contrast phase CT) are shown with PCA plot using self-supervised contrastive loss (SSCL) and radiological conditional contrastive loss. With AGCL, the organ representation can be well separated into specific modal clusters, while the separation of clusters is not well distinguished by using SSCL.} \label{PCA}
\end{figure*}

\begin{itemize}
    \item $Coarse\:Segmentation\:Module$, we use RAP-Net as the  coarse segmentation backbone and generate multiple objects segmentation mapping for each image \cite{lee2021rap}. The segmentation mapping is then converted to a binary object-corresponding attention map. $P_i$ is randomly sampled for each object within the attention map bounded region. Further processing details in patch extraction are provided in the Supplementary Material.
    \item $Pairwise\:Data\:Augmentation\:Module$, $P_{Aug}(\cdot)$, given a batch of multi-modal mixed query patches, pairwise copies with multiple views are generated $\widetilde{a}_i= P_{Aug}({a}_i)$ with random data augmentations from $a_i$. The augmentation module consists of random cropping, rotation (random from $-{30^{\circ}}$ to $+{30^{\circ}}$), and random scaling (width: 0.3, length: 0.7). 
    \item $Encoder\:Network$, $E(\cdot)$, ResNet-like network architectures are used as the encoder backbone and generate pairwise feature representation vector $\widetilde{z}_i= E(\widetilde{a}_i), \widetilde{z}_i\in{R^{O_E}}$, where $O_E$ is the size of the output space. The goal of the encoder is to learn multiple discrete clusters for corresponding categories. The feature representation is mapped as a point to the hypersphere with a radius of $1/T$, where $T$ is a hyperparameter indicating temperature scaling to control the weighting on the positive/negative pairs. The contrastive pre-trained encoder then has its weights frozen and the representation vector $z_i = E{\{{a}_i\}}$ is computed for fine-tuning segmentation task as the second stage.
    \item $Projection\;Network$, $P(\cdot)$, we follow a similar structure to \cite{chen2020simple} use a multi-layer perceptron with an output vector size of $O_E = 256$. We compute the distances between each point in the projection space and meet our goal of defining independent embeddings with corresponding modality and organ constraints. The projection network is discarded after the encoder section is well trained. Sec.III B gives more details on the theory behind our proposed contrastive loss function. 
    \item $Decoder\:Network$, $D(\cdot)$, A contrastive representation decoder with atrous spatial pyramid pooling (ASPP) module is utilized to generate binary segmentation predictions. The segmentation loss for the medical image domain $\mathcal{L}_{seg}$ is computed with the binary organ-corresponding label $ y_i$ as the following:
    \begin{equation}
        \mathcal{L}_{seg} = \sum{\bigg(1-2\cdot \frac{s_i\cdot{y_{i}}}{s_i+{y_{i}}}\bigg)}
    \end{equation}
    where $s_i = D(z_i), s_i\in{R^{b\times 2\times h\times w}}$, $y_i$ is the binary one-hot ground truth, $b$ is the batch size, $h$ and $w$ are the height and width of the image patch respectively.
\end{itemize}

\newcommand{\tabincell}[2]{\begin{tabular}{@{}#1@{}}#2\end{tabular}}
\begin{table*}[htb]
    \centering
    \caption{Comparison of the fully-supervised, unsupervised, semi-supervised and partially supervised state-of-the-art methods on the 2015 MICCAI BTCV challenge leaderboard. (We show 8 main organs Dice scores due to limited space, $\ast$: fully-supervised approach, $\star$: semi-supervised approach, $\bigtriangleup$: partially supervised approach.)}
    \begin{tabular}{*{1}{l}|*{8}{c}|*{3}{c}}
        \toprule 
         Method & Spleen & R.Kid & L.Kid & Gall. & Eso. & Liver & Aorta & IVC &\tabincell{c}{Average\\ Dice} &\tabincell{c}{Mean Surface\\ Distance} &\tabincell{c}{Hausdorff\\ Distance} \\
        
         \midrule
         $Cicek\:et\:al.\:\ast$\cite{cciccek20163d} & 0.906 & 0.857 & 0.899 & 0.644 & 0.684 & 0.937 & 0.886 & 0.808 & 0.784 & 2.339 & 15.928\\
         $Roth\:et\:al.\:\ast$\cite{roth2018multi} & 0.935 & 0.887 & 0.944 & 0.780 & 0.712 & 0.953 & 0.880 & 0.804 & 0.816 & 2.018 & 17.982\\
         $Heinrich\:et\:al.$\cite{heinrich2015multi} & 0.920 & 0.894 & 0.915 & 0.604 & 0.692 & 0.948 & 0.857 & 0.828 & 0.790 & 2.262 & 25.504 \\
         $Pawlowski\:et\:al.\:\ast$\cite{pawlowski2017dltk} & 0.939 & 0.895 & 0.915 & 0.711 & 0.743 & 0.962 & 0.891 & 0.826 & 0.815 & 1.861 & 62.872\\
         $Zhu\:et\:al.\:\ast$\cite{zhu2019multi} & 0.935 & 0.886 & 0.944 & 0.764 & 0.714 & 0.942 & 0.879 & 0.803 & 0.814 & 1.692 & 18.201\\
         $Lee\:et\:al.\:\ast$\cite{lee2021rap} & 0.959 & 0.920 & 0.945 & 0.768 & 0.783 & 0.962 & 0.910 & 0.847 & 0.842 & 1.501 & 16.433\\
         $Zhou\:et\:al.\:\bigtriangleup$\cite{Zhou_2019_ICCV} & 0.968 & 0.920 & 0.953 & 0.729 & 0.790 & 0.974 & 0.925 & 0.847 & 0.850 & 1.450 & 18.468 \\
         $Chaitanya\:et\:al.\:\star$\cite{chaitanya2020contrastive} & 0.956 & 0.935 & 0.946 & 0.920 & 0.854 & 0.970 & 0.915 & 0.893 & 0.874 & 1.236 & 15.281 \\
         $Wang\:et\:al.\:\star$\cite{wang2021dense} & 0.963 & 0.939 & 0.900 & 0.815 & 0.838 & 0.976 & 0.922 & 0.907 & 0.882 & 1.303 & 14.759 \\
         $Khosla\:et\:al.\:\ast$\cite{khosla2020supervised} & 0.959 & 0.939 & 0.947 & \textbf{0.932} & 0.867 & 0.978 & 0.922 & 0.911 & 0.907 & 0.978 & 14.136 \\
         \midrule
         Ours (SSCL)$\:\star$ & 0.953 & 0.922 & 0.930 & 0.830 & 0.822 & 0.972 & 0.899 & 0.874 & 0.863 & 1.899 & 17.073\\
         \textbf{Ours (AGCL)}$\:\ast$ & \textbf{0.971} & \textbf{0.955} & \textbf{0.963} & 0.910 & \textbf{0.886} & \textbf{0.984} & \textbf{0.941} & \textbf{0.932} & \textbf{0.923} & \textbf{0.932} & \textbf{13.024} \\ \bottomrule
    \end{tabular}
    \label{baselines_compare}
\end{table*}

\begin{figure*}[htb]
\centering
\includegraphics[width=0.8\textwidth]{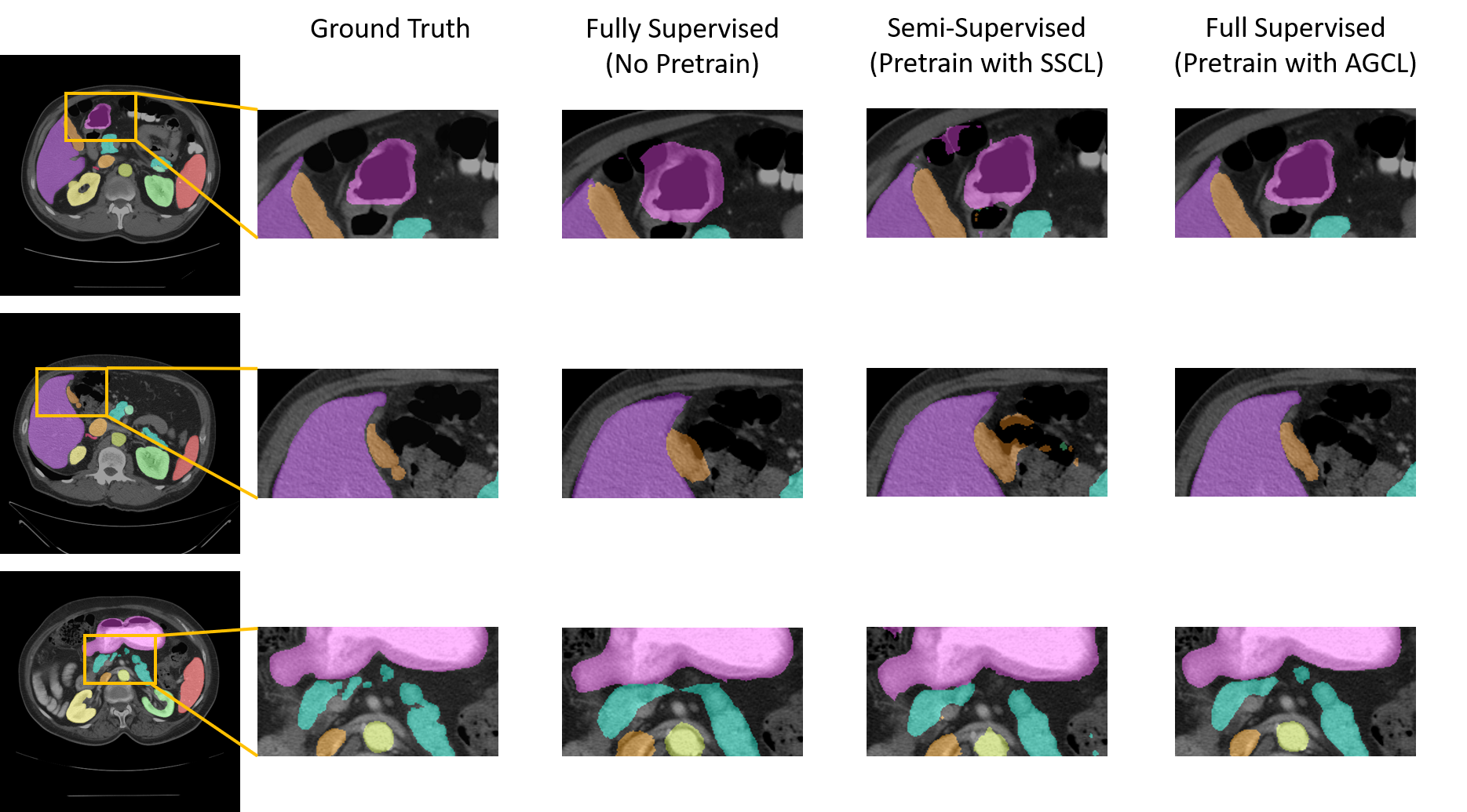}
\caption{Visualization of segmentations with different state-of-the-arts pre-training strategies demonstrate incremental improvement on segmentation performance. AGCL demonstrates smooth boundaries and accurate morphological information between neighboring organs, while SSCL is prone to identifying false positive in neighboring areas.}
\label{qualitative_figure}
\end{figure*}

\subsection{Multi-Class Conditional Contrastive Loss}
We adapt the contrastive loss for learning independent embeddings. We initially compute the pairwise augmented samples for training in a total of $2N$ pairs, $\{\widetilde{a}_i, l_i\}_{i=1,...,2N}$, where $\widetilde{a}_{2k}$ and $\widetilde{a}_{2k-1}$ are the two randomly augmented versions from the original image patch with different views as $k = 1...N$, and $l_i=\{m_i\in1...M, o_i\in1...O\}$ is the corresponding modal-target information, where $M$ and $O$ represents the number of modalities and the number of semantic targets. As shown in Fig. 3, we find that some of the organ representations can not be well separated with the use of self-supervised contrastive loss defined as the following:
\begin{equation}
    \mathcal{L}_{self} = -\sum_{k=1}^{2N} \log \frac{{\exp(\widetilde{z}_{k} \cdot \widetilde{z}_{p(k)}/\mathcal{T})}}{{\sum_{j\in{J(k)}}} \exp({\widetilde{z}_{k} \cdot \widetilde{z}_{j}/\mathcal{T})}}
    \label{Eq.2}
\end{equation}
where  $\widetilde{z}_{k}$ and $\widetilde{z}_{p(k)}$ are the pairwise feature representation vectors generated from $\widetilde{z}_{k}, \widetilde{z}_{p(k)} = P(E(a_{2k}, a_{2k-1})$. The index $k$ represents the sample of anchor and index $p(k)$ represents the corresponding positive. With SSCL, only one positive pair $\widetilde{z}_{k}$ and $\widetilde{z}_{p(k)}$ is classified among all pairs within a minibatch and the other $2(k-1)$ representation vectors $\widetilde{z}_{j}$ are classified as the negative pairs. As such, SSCL is limited in classifying organ representations in general. For instance, organ query patches from another subject that contain the same organ may be indicated as negative pairs and lead to inaccurate representations in the latent space. Therefore, we extend the contrastive loss with conditional constraints to increase the number of positive pairs by defining representation with the same organ and modality as positive pairs across all subjects images in a minibatch, generalizing Eq.\ref{Eq.2} with multi-class supervision:
\begin{equation}
    \mathcal{L}_{MT} = \sum_{k=1}^{2N} \frac{-1}{|L(k)|} \sum_{l\in L(k)} \log \frac{{\exp(\widetilde{z}_{k} \cdot \widetilde{z}_{l}/\mathcal{T})}}{{\sum_{j\in{J(k)}}} \exp({\widetilde{z}_{k} \cdot \widetilde{z}_{j}/\mathcal{T})}} \\
    \label{Eq.3}
\end{equation}

\noindent Here, $ L(k) \equiv \{l \in J(k): \{m_{l}, o_{l}\} = \{m_{i}, o_{i}\}\}$ and $|L(k)|$ is the cardinality of the representation vector corresponding to modalities and semantic targets respectively. Motivated by \cite{khosla2020supervised}, each embedding is defined as the corresponding modality and object. The integration of multi-classes context increases the number of positives and increases the encoding ability towards corresponding embeddings. The organ representation in each modality is well separable as independent clusters (Fig. 3).

\section{Experiments}
\textbf{Datasets}: To evaluate our proposed learning approach, one in-house research cohorts and two publicly available datasets in medical imaging and natural imaging domain are used. \\
\textbf{[$\mathbf{I}$]} \textbf{MICCAI 2015 Challenge Beyond The Cranial Vault (BTCV) dataset} is comprised of 100 de-identified 3D contrast-enhanced CT scans with 7968 axial slices in total. 20 scans are publicly available for the testing phase in the MICCAI 2015 BTCV challenge. For each scan, 12 organ anatomical structures are well-annotated, including spleen, right kidney, left kidney, gallbladder, esophagus, liver, stomach, aorta, inferior vena cava (IVC), portal splenic vein (PSV), pancreas and right adrenal gland. Each volume consists of $47\sim133$ slices of $512\times512$ pixels, with resolution of $([0.54\sim0.98]\times[0.54\sim0.98]\times[2.5\sim7.0])mm^{3}$. \\ \textbf{[$\mathbf{II}$]} \textbf{Non-contrast clinical cohort} is retrieved in de-identified form from ImageVU database of Vanderbilt University Medical Center. It consists of 56 3D CT scans with 3687 axial slices and expert refined annotations for the same 12 organs as the MICCAI 2015 BTCV challenge datasets. Each volume consists of $49\sim174$ slices of $512\times512$ pixels, with resolution of $([0.64\sim0.98]\times[0.64\sim0.98]\times[1.5\sim5.0])mm^{3}$. \\ \textbf{[$\mathbf{III}$]} \textbf{PASCAL VOC 2012} dataset consists of 10582 training images, 1449 validation images and 456 testing images with pixel-level annotations of 20 classes and one background.

\begin{figure*}[htb]
    \centering
    \includegraphics[width=0.9\textwidth]{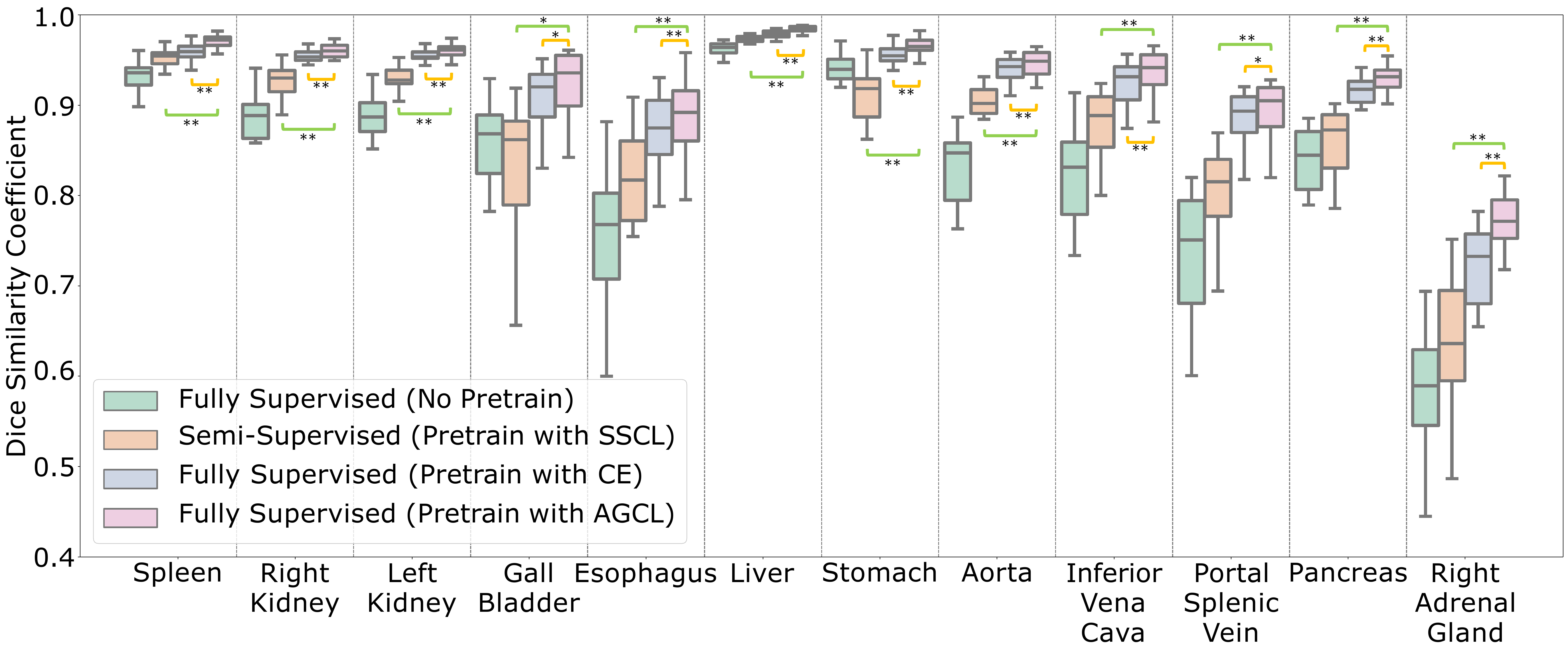}
    \caption{AGCL outperforms the current state-of-the-art pre-training methods with SSCL and classification pre-training with CE across all organs. (*: p$<$0.05, **: p$<$0.01, with Wilcoxon signed-rank test).}
    \label{ablation_resnet50}
\end{figure*}

\begin{table*}[htb]
    \centering
    \caption{Ablation studies of segmentation performance in various network backbones of the BTCV testing cohort.}
    \begin{tabular}{{c c|}*{12}{c}}
        \toprule\toprule
        Encoder & Pretrain & Spleen & R.Kid & L.Kid & Gall. & Eso. & Liver & Stomach & Aorta & IVC & PSV & Pancreas & R.A \\
        \midrule
        ResNet50 & $\times$ & 0.932 & 0.877 & 0.887 & 0.860 & 0.761 & 0.962 & 0.941 & 0.832 & 0.815 & 0.735 & 0.833 & 0.587 \\
        ResNet50 & SSCL & 0.953 & 0.922 & 0.930 & 0.842 & 0.822 & 0.972 & 0.907 & 0.899 & 0.874 & 0.800 & 0.854 & 0.625 \\
        ResNet50 & CE & 0.959 & 0.948 & 0.957 & 0.890 & 0.868 & 0.978 & 0.956 & 0.935 & 0.919 & 0.884 & 0.903 & 0.725 \\
        ResNet50 & AGCL & \textbf{0.971} & \textbf{0.955} & \textbf{0.963} & \textbf{0.910} & \textbf{0.886} & \textbf{0.984} & \textbf{0.965} & \textbf{0.941} & \textbf{0.932} & \textbf{0.893} & \textbf{0.917} & \textbf{0.769} \\
        \midrule
        ResNet101 & $\times$ & 0.909 & 0.849 & 0.864 & 0.859 & 0.694 & 0.953 & 0.909 & 0.770 & 0.767 & 0.713 & 0.814 & 0.526 \\
        ResNet101 & SSCL & 0.950 & 0.928 & 0.935 & 0.805 & 0.792 & 0.969 & 0.900 & 0.905 & 0.877 & 0.800 & 0.846 & 0.602 \\
        ResNet101 & CE & 0.960 & 0.933 & 0.945 & 0.887 & 0.822 & 0.975 & 0.952 & 0.920 & 0.901 & 0.834 & 0.877 & 0.670\\
        ResNet101 & AGCL & \textbf{0.965} & \textbf{0.948} & \textbf{0.954} & \textbf{0.901} & \textbf{0.875} & \textbf{0.981} & \textbf{0.962} & \textbf{0.930} & \textbf{0.917} & \textbf{0.876} & \textbf{0.902} & \textbf{0.748} \\
        \bottomrule\bottomrule
    \end{tabular}
    \label{ablation_resnet}
\end{table*}

We evaluate the segmentation performance with Dice similarity coefficient and mean IoU on a number of segmentation benchmarks for medical and natural image domain respectively, including the testing phase of the BTCV dataset, testing cohort of the non-contrast clinical cohort and the validation set of PASCAL VOC 2012 dataset for multi-classes segmentation. We performed ablation studies with the variation of hyperparameters and the training strategies for encoder network structure in the first stage. For the encoder network, we evaluated with two common backbone architectures both in medical imaging domain and natural image domain for segmentation: Deeplabv3+ with ResNet-50 and ResNet-101 encoder. The normalized activation of the final pooling layer with $D_E = 2048$ are used as the distinctive feature representation vector. We provide more preprocessing, training and implementation details in the Supplementary Material.

\subsection{Segmentation Performance}
We first compare the proposed AGCL with a series of state-of-the-art approaches including 1) fully supervised approaches (training on ground-truth labeled data only), 2) a partially-supervised approach (training on one contrast phase dataset, and another with partial labels), and 3) a semi-supervised approach (contrastive learning with self-supervised contrastive loss). As shown in Table I, the semi-supervised approach demonstrates significant improvement followed by the partial-supervision and full-supervision approaches. $Chaitanya\:et\:al.$ integrate the self-supervised contrastive loss across global to local scale and demonstrate significant improvement across organs. $Khosla\:et\:al.$ provide an additional single class label to address the correspondence on embeddings, which outperforms all current approaches in supervised and semi-supervised settings. By further adding modality and anatomical information as conditional constraints, AGCL achieves the best performance among all state-of-the-arts with a mean Dice score of 0.923. The additional gains demonstrate that our use of supplemental imaging information allows for recognition of more positive pairs. Beyond the medical imaging dataset, we performed experiments on the natural image dataset PASCAL VOC 2012 for multi-class segmentation in a single model. In Table III, AGCL demonstrates substantial improvement on segmentation performance when compared against the current supervised state-of-the-arts. With the use of cross-entropy loss, AGCL outperforms the original DeepLabv3+ methods by 2.75 percent. Interestingly, the improvement is comparatively less than that with cross-entropy by using the RMI loss. 

\begin{figure*}[htb]
    \centering
    \begin{subfigure}[b]{0.3\textwidth}
    \includegraphics[width=\textwidth]{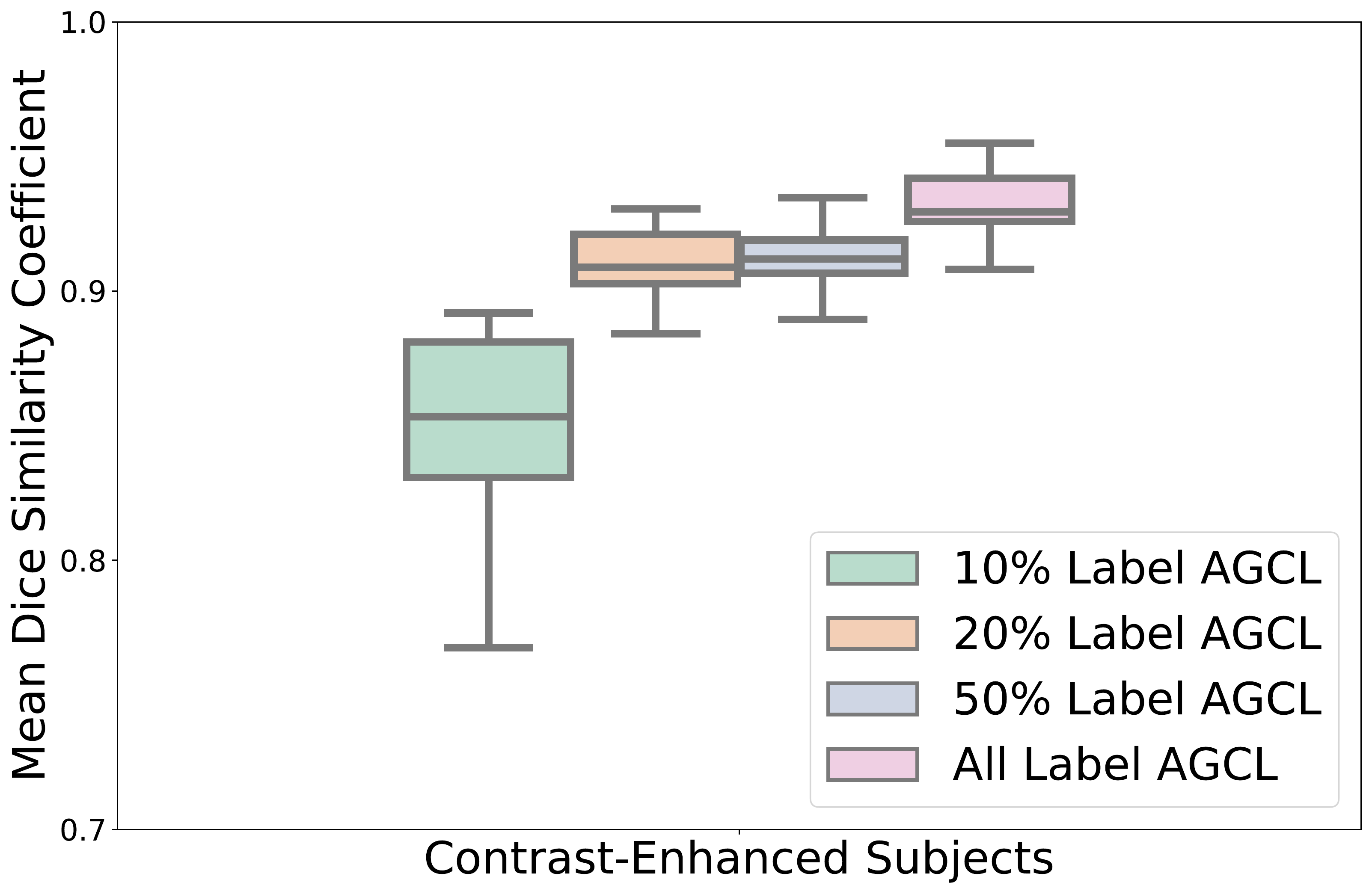}
    \caption{}
    \end{subfigure}
    \begin{subfigure}[b]{0.3\textwidth}
    \includegraphics[width=0.95\textwidth]{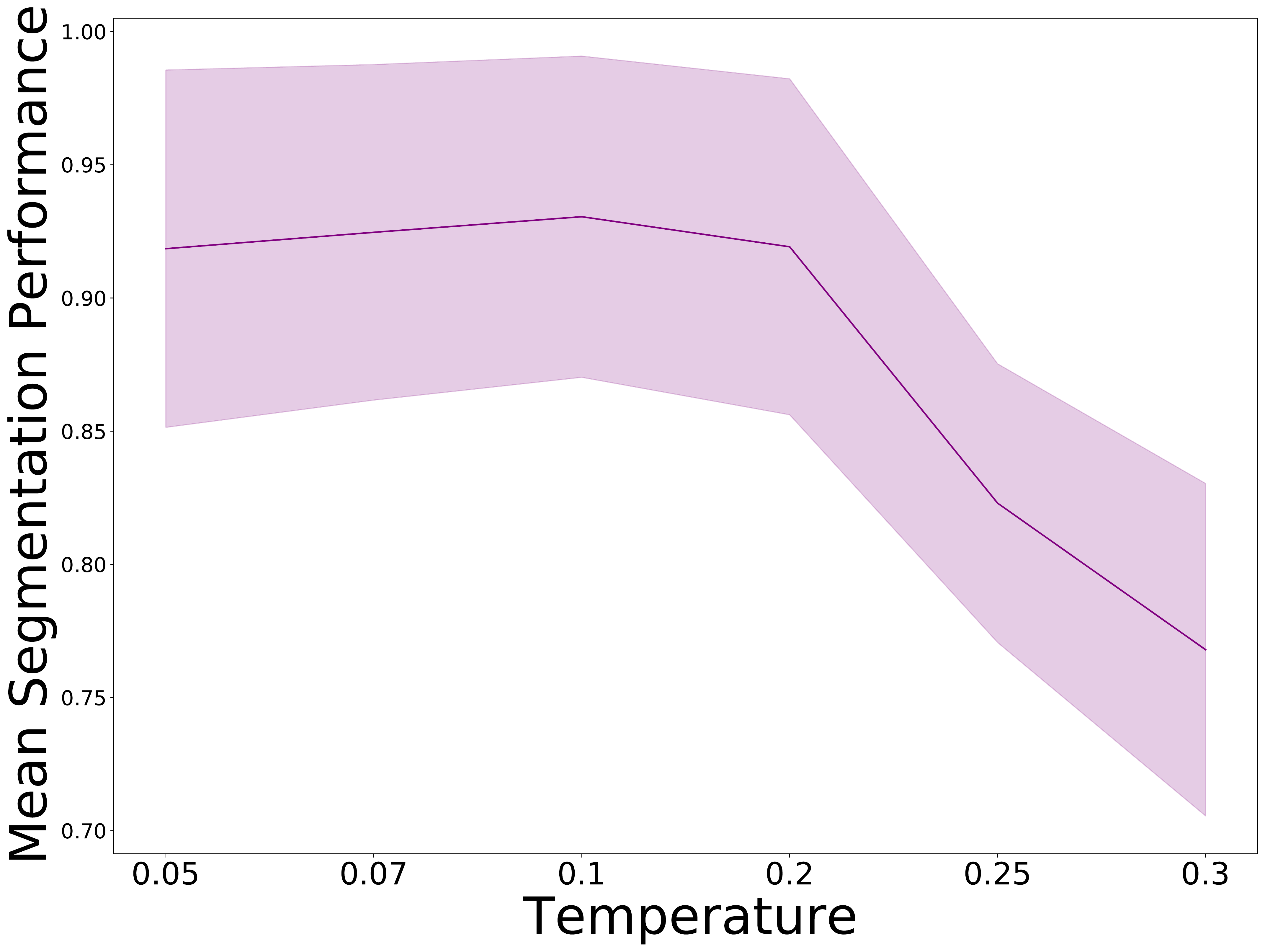}
    \caption{}
    \end{subfigure}
    \begin{subfigure}[b]{0.3\textwidth}
    \includegraphics[width=\textwidth]{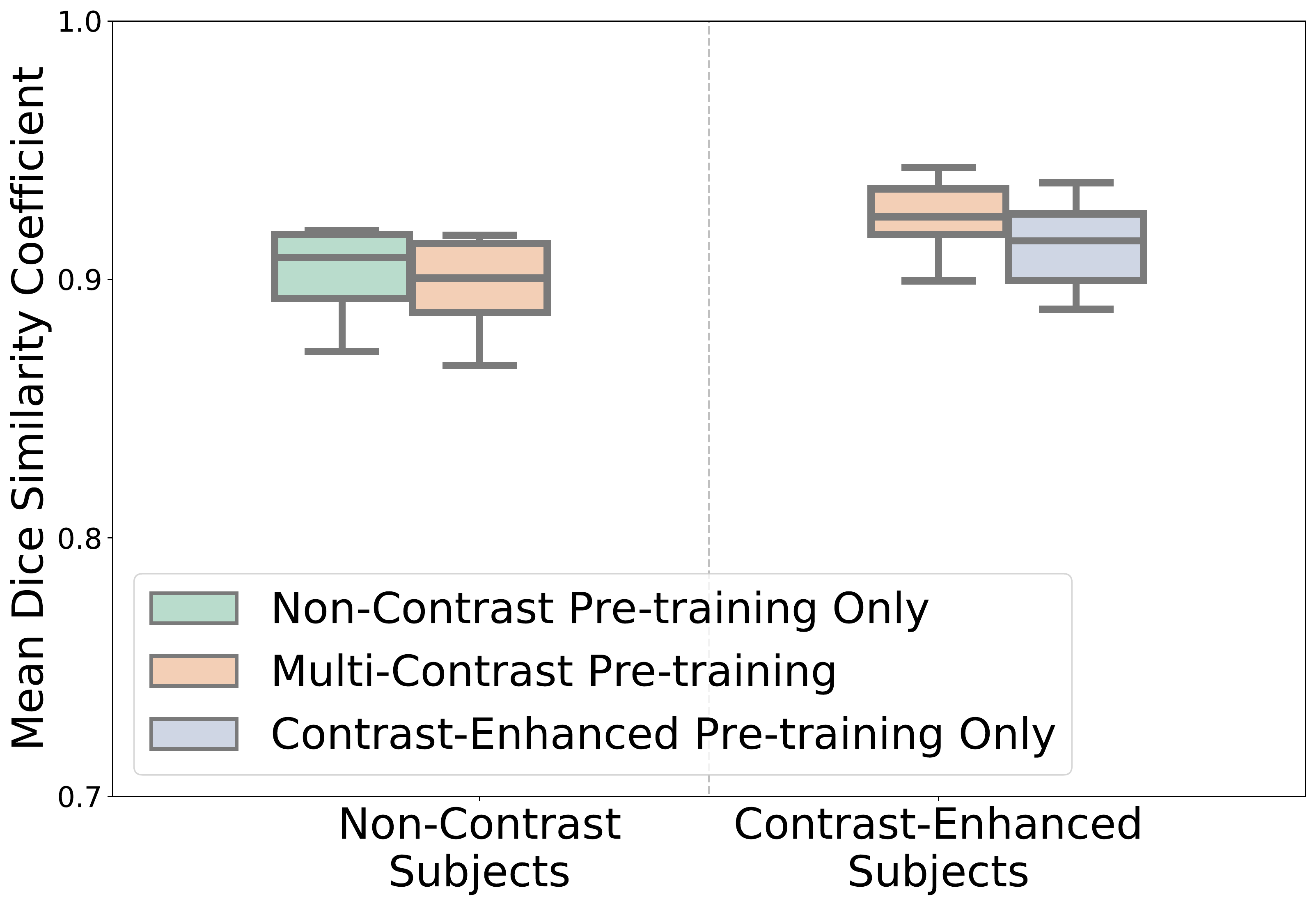}
    \caption{}
    \end{subfigure}
\caption{\textbf{a)} The segmentation performance gradually improved with the additional quantities of image-level labels for AGCL. \textbf{b)} Ablation studies of temperature scaling the distance between positive/negative pairs demonstrates that the segmentation performance is best optimized when $T=0.1$. \textbf{c)} The segmentation performance on contrast-enhanced datasets is better with multi-modal training, while non-contrast performance improves with single modal training only.}
\end{figure*}


\subsection{Ablation Study for AGCL}
\textbf{Comparing with first stage training approaches:} We perform evaluation among the three supervised and self-supervised approaches in the first stage: 1) training with self-supervised contrastive loss (SSCL), 2) training with cross-entropy (CE) loss as classification tasks, and 3) random initialization (RI) without any contrastive learning on both ResNet-50 and ResNet-101 encoder backbone. \\
\indent As shown in Table II and Fig. 5, SSCL improves the segmentation performance over RI by 3.12\%, which is expected because RI considers no constraints in the lower-dimensional space and relies on the decoder ability for downstream tasks. With the supervision of modality and anatomical information, the supervised CE strategies significantly boost the segmentation performance by 5.93\%. It demonstrates that a good definition of the latent space in encoder can help address the corresponding representation for each semantic target and starts to achieve more favorable with the segmentation task. Eventually, AGCL surpasses CE by 1.32\% in mean Dice and showed that the multi-class conditional contrastive loss can provide a better definition on separating embeddings than CE. \\
\indent To further evaluate the segmentation using different contrastive learning approaches, the qualitative representation of the segmentation prediction with each training method is demonstrated on Fig. 4 comparing with the ground truth label. With SSCL, the boundary of the segmentation is significantly smoother than that of with RI. However, we found that additional segmentation is performed near the neighboring structures. The similar intensity range and morphological appearance may lead to the instability of representation extraction from SSCL. With the additional constraints by AGCL, the boundary information between neighboring organs are clearly defined and the segmentation quality is comparable to the ground truth label.

\textbf{Comparing with temperature variability:} We experimented with the variation of temperature to investigate the optimal effect towards the segmentation performance. Fig. 6(b) demonstrate the effect of temperature on the multi-organ segmentation across all subjects in the BTCV testing dataset. We found that low temperature achieves better performance than high temperature, as the radius of the hypersphere defined in the latent space is inversely proportional to the temperature scaling and increase the difficulty of finding positive samples with the decrease of radius.

\textbf{Comparing with single/multiple modal contrastive learning}
The segmentation performance is evaluated with single modality and with multi-modality contrastive learning respectively. From Fig. 6(c), a better segmentation performance for contrast-enhanced dataset is achieved by contrastive learning with multi-modality images. Interestingly, we observed that the segmentation performance of non-contrast imaging is improved to a small extent with non-contrast modal pre-training only.

\textbf{Comparing with reduced label for AGCL:} In Fig. 6(a), we performed AGCL with the variation of label quantity and compare the segmentation performance by leveraging the amount of label information. We observed that model has the best performance with fully labeled input. A significant improvement is shown with 20\% labels for AGCL comparing to that with 10\% labels, while an improvement to a small extent is demonstrated by using 50\% label for AGCL.

\subsection{Limitations}
Although AGCL tackles current challenges of integrating contrastive learning into multi-object segmentation, limitations still exist in the process of AGCL. One limitation is the dependency on the coarse segmentation quality. As 2D patches are extracted with the attention information in each slice, patches without corresponding organ regions may also be possible
\begin{wrapfigure}{r}{0.5\textwidth}
\centering
\begin{minipage}{\linewidth}
    \centering
    \captionof{table}{Average segmentation performance in Mean IoU across all classes on PASCAL VOC 2012 validation set using single model with the ResNet-50 backbone.}
    \begin{tabular}{cccc} 
    \toprule
    \multirow{2}*{Architecture} & \multicolumn{3}{c}{Loss}\\\cmidrule{2-4}
    \quad  &  CE & AAF \cite{ke2018adaptive} & RMI \cite{zhao2019region} \\\midrule
    DeepLabv3 \cite{chen2017deeplab} & 0.7311 & 0.7234 & 0.7539 \\
    DeepLabv3+ \cite{chen2017rethinking} & 0.7204 & 0.7268 & 0.7681 \\
    PSP-Net \cite{zhao2017pyramid} & 0.7301 & 0.7256 & 0.7551 \\
    \textbf{AGCL (Ours)} & \textbf{0.7402} & \textbf{0.7456} & \textbf{0.7722} \\\bottomrule
    \end{tabular}
    \label{pascal_baselines}
\end{minipage}
\end{wrapfigure}
\\
\\
\\
\\
\\
\\
\\
\\
\\
\\
\\
\\
\\
to extract due to inaccurate segmentation. Incorrect label definition inputs may bring into contrastive learning process. Another limitation is performing contrastive learning in object-centric setting. We aim to innovate segmentation pipeline with contrastive learning as an end-to-end setting in our future work. 

\section{Conclusion}
Performing robust multi-object semantic segmentation using deep learning remains a persistent challenge. In this work, we propose a novel semantic-aware contrastive framework that extends self-supervised contrastive loss and integrates attention guidance from coarse segmentation. Our proposed method leads to a significant gain in segmentation performance on two CT datasets and one natural image dataset.

\ifCLASSOPTIONcaptionsoff
  \newpage
\fi



\bibliographystyle{IEEEtran}
\bibliography{main}
%







\end{document}